\documentclass[11pt, a4paper, logo, copyright]{googledeepmind}

\usepackage[dvipsnames]{xcolor}         
\usepackage{amsmath} 
\usepackage{mathtools}
\mathtoolsset{showonlyrefs}
\usepackage{amssymb}
\usepackage{bm}

\usepackage{thm-restate}
\usepackage{algorithm}
\usepackage[noend]{algpseudocode}

\usepackage{enumitem}

\usepackage[authoryear, compress, round]{natbib}
\def\d{{\bm{d}}}

\def\s{{\bm{s}}}
\def\u{{\bm{u}}}
\def\w{{\bm{w}}}
\def\x{{\bm{x}}}
\def\y{{\bm{y}}}

\def\softargmax{{\mathrm{softargmax}}}
\def\lse{{\mathrm{LSE}}}

\def\ymax{{\widehat{\y}^{\max}}}

\def\ebm{{\mathrm{EBM}}}
\def\ar{{\mathrm{AR}}}
\def\greedy{{\mathrm{greedy}}}
\def\sampling{{\mathrm{sampling}}}
\def\mm{{\mathrm{MM}}}

\def\cM{{\mathcal{M}}}
\def\cP{{\mathcal{P}}}

\def\cV{{\mathcal{V}}}
\def\cW{{\mathcal{W}}}
\def\cX{{\mathcal{X}}}
\def\cY{{\mathcal{Y}}}

\def\wty{{\widetilde{y}}}
\def\wtyv{{\widetilde{\y}}}

\def\RR{{\mathbb{R}}}

\def\EE{{\mathbb{E}}}

\def\yhat{\widehat{\y}}

\DeclareMathOperator*{\argmax}{argmax}

\def\iid{{i.i.d.\ }}
\def\aka{{a.k.a.\ }}
\def\wrt{{w.r.t.\ }}

\usepackage{mdframed}
\definecolor{theoremcolor}{rgb}{255, 255, 255}
\mdfsetup{
    backgroundcolor=theoremcolor,
    linewidth=1pt,
    leftline=true,
    topline=false,
    rightline=false,
    bottomline=false 
}

\newmdtheoremenv{definition}{Definition}
\newmdtheoremenv{proposition}{Proposition}
\newmdtheoremenv{corollary}{Corollary}
\newmdtheoremenv{theorem}{Theorem}
\newmdtheoremenv{lemma}{Lemma}
\newmdtheoremenv{example}{Example}
\newmdtheoremenv{remark}{Remark}

\title{On the Difficulty of Sequence-Level Supervised Learning for Autoregressive Language Models}

\correspondingauthor{mblondel@google.com}

\author[1]{Mathieu Blondel}
\author[1]{Micha\"el E. Sander}
\author[1]{Germain Vivier-Ardisson}
\author[1]{Tianlin Liu}
\author[1]{Vincent Roulet}

\affil[1]{Google DeepMind}

\begin{abstract}
During the supervised learning phase, autoregressive language models are
typically trained by minimizing the negative log-likelihood of the training
data, which decomposes as a sum of token-level cross-entropies via teacher
forcing. 
This objective is simple and efficient but potentially suffers from so-called
exposure bias: a mismatch between the way the model is trained and the way it is
used at inference time.  Can we instead design sequence-level loss functions
that incorporate inference-time decoding procedures, such as greedy decoding or
autoregressive sampling?
We show that two natural attempts lead to loss functions that are fundamentally
flawed: unbounded below.  
We further show that these two flawed loss functions are intimately connected:
using autoregressive Gumbel tricks, the sampling based loss is the expected
greedy loss under Gumbel perturbations.
Since the pathology stems from the dependence of the decoded sequence on the
model parameters, we analyze the update direction obtained by applying a stop
gradient on the sampling procedure, 
leading to a moment matching direction.  While it is not the gradient
of any loss function, we prove that it defines a valid descent direction in
function space for the autoregressive negative log-likelihood. 
Despite these theoretical insights, the moment matching direction suffers from
high variance. We conclude that teacher forcing may well be ``all you need''
for supervised learning of autoregressive models.  
\end{abstract}

\begin{document}

\maketitle

\section{Introduction}

During the supervised learning phase, autoregressive (AR) language models are
typically trained by minimizing the negative log-likelihood of the training
data, which decomposes as a sum of \textbf{token-level} cross-entropies.  During
training, each token is predicted given the ground-truth prefix, a technique
known as \textbf{teacher forcing} \citep{williams1989learning}. 

While simple and computationally efficient
(a single forward pass, no sampling), teacher forcing
potentially suffers from \textbf{exposure bias}
\citep{ranzato2015sequence, bengio2015scheduled}: at training time, the model
conditions on ground-truth prefixes, but at inference time it conditions on its
own previous predictions, leading to error accumulation.

These limitations have motivated work on \textbf{sequence-level} training
objectives.  Existing efforts include
policy-gradient \citep{ranzato2015sequence, rennie2017self} and
minimum risk training \citep{shen2016minimum, norouzi2016reward,
edunov2018classical} methods. 
However, these methods all rely on external reward signals.
In the structured prediction literature, 
designing loss functions for structured outputs has been studied extensively.
The perceptron loss \citep{collins2002discriminative} requires an
\textbf{exact} maximization oracle, while negative log-likelihood training of
conditional random fields and more generally energy-based models
\citep{lecun2006tutorial} requires a log-sum-exp (log-partition)
oracle or a MCMC sampler. Unfortunately, such oracles or samplers
are intractable over the space of sequences.

Can we instead design sequence-level loss functions that incorporate 
inference-time decoding procedures, such as greedy decoding or
autoregressive sampling? In this paper, we make the following contributions.
\begin{itemize}[itemsep=6pt]

\item After reviewing some background (Section \ref{sec:background}), 
we show that the sequence-level loss functions associated with greedy decoding
and autoregressive sampling are \textbf{unbounded below}
(Section~\ref{sec:flawed}).
These loss functions can therefore not be used for learning, as the loss would
diverge to $-\infty$.

\item We connect greedy decoding and sampling through \textbf{autoregressive
    Gumbel tricks}, showing that the sampling based loss is the expected greedy
    loss under Gumbel perturbations
    (Section~\ref{sec:autoregressive_gumbel_tricks}).

\item Since the pathology of these loss functions stems from the dependence of
    the decoded sequence on the model parameters, 
    we then analyze the \textbf{moment matching direction} obtained via a stop
gradient on the sampling procedure: it is a surrogate gradient (not the gradient
of any loss), yet a valid descent direction for the AR negative log-likelihood
(Section~\ref{sec:moment_matching}) in function space.
Unfortunately, this direction suffers from high variance.

\item We conclude that teacher forcing may well be ``all you need'' for
supervised learning of AR models.

\end{itemize}

\paragraph{Notation.}

We follow the same notation as \citet{blondel2025autoregressive}.
We define the space of input sequences as $\cX$ and the space of output
sequences as $\cY$.
For notational simplicity, we assume $\cY \coloneqq \cV^T$, where $T$ is a fixed
length and $\cV \coloneqq \{1,\dots,V\}$. For a detailed discussion on
variable-length sequence handling, see \citet{blondel2025autoregressive}.
Given a sequence $\y = (y_1, \dots, y_T)$, $\y_{<t}$ denotes the sequence $(y_1,
\dots, y_{t-1})$ and for convenience $\y_{<1}$ denotes the empty sequence.
We use $\oplus$ to denote sequence concatenation.
We denote random variables with capital letters, for example, $Y$.
We define the space of probability distributions over $\cY$ as $\cP(\cY)$.
We denote \texttt{stop\_gradient} the operation that keeps forward
computations in automatic differentiation unchanged while zeroing out
the backpropagation of gradients in the backward pass,
see e.g., \citet{edpbook} for more details.

\section{Background}
\label{sec:background}

In this section, we review energy-based models (EBMs) and autoregressive (AR)
models, emphasizing their shared structure and the loss functions associated
with them.

\subsection{Energy-based models}

We are given a function $R(\x, \y)$,
typically implemented as a bidirectional Transformer,
which assigns a real-valued ``affinity'' score to a pair $(\x, \y)$.
We will also write $R_\w$ to indicate the dependence of $R$ on some parameters
$\w \in \cW$ and we suppose throughout this paper that $R_\w$ is differentiable
in $\w$.
We then define probabilistic energy-based models (EBMs) as
\begin{equation}
p^\ebm_R(\y|\x) 
\coloneqq \frac{\exp(R(\x, \y))}{\sum_{\y' \in \cY} \exp(R(\x, \y'))}.
\end{equation}
We can see probabilistic EBMs as specifying a \textbf{sequence-level}
Gibbs distribution (\aka Boltzmann distribution) parameterized by $R$.

\paragraph{Sampling.}

Drawing \iid sequence samples
\begin{equation}
Y \sim p^\ebm_R(\cdot|\x)
\end{equation}
typically requires Markov-chain Monte-Carlo (MCMC) methods, such as Gibbs
sampling. These methods circumvent the intractable computation of the
normalization factor in the probability definition.
However, such samplers are typically inherently sequential and
difficult to parallelize. Moreover, since they are never run to convergence in
practice, they do not generate truly \iid samples.

\paragraph{Sequence-level log-sum-exp and associated loss.}

The cumulant is the sequence-level log-sum-exp (\aka log-partition function):
\begin{equation}
A^\lse_R(\x) 
\coloneqq \underset{\y \in \cY}{\lse} ~ R(\x, \y)
\coloneqq \log \sum_{\y \in \cY} \exp(R(\x, \y)).
\end{equation}
The brute force computation cost is $O(V^T)$. 

The associated loss is the negative log-likelihood:
\begin{equation}\label{eq:lebm}
L^\ebm_R(\x, \y) 
\coloneqq -\log p^\ebm_R(\y|\x) 
= A^\lse_R(\x) - R(\x, \y).
\end{equation}
The loss is convex in $R$ and satisfies the property
\begin{equation}
L^\ebm_R(\x, \y) = 0 \iff p^\ebm_R(\y|\x) = 1.
\end{equation}
The gradient \wrt parameters $\w \in \cW$ is \citep{song2021train}
\begin{equation}
\nabla_\w L^\ebm_{R_\w}(\x, \y) 
= \EE_{Y \sim p^\ebm_{R_\w}(\cdot|\x)}\left[\nabla_\w R_\w(\x, Y)\right] -
\nabla_\w R_\w(\x, \y).
\end{equation}
Unfortunately, this loss function requires access either to a log-sum-exp
oracle, also known as marginal inference oracle \citep{wainwright2008graphical},
or being able to sample from $p_{R_\w}^\ebm$.
Both are intractable in general over the space of sequences $\cY$.

\paragraph{Exact maximization oracle and associated loss.}

The \textbf{mode} of the distribution is
\begin{equation}
\widehat{\y}^{\max}_R(\x)
\coloneqq \argmax_{\y \in \cY} ~ p^\ebm_R(\y|\x)
= \argmax_{\y \in \cY} ~ R(\x, \y) \in \cY.
\end{equation}
We can also define the value of the most plausible sequence:
\begin{equation}
A_R^{\max}(\x) \coloneqq \max_{\y \in \cY} ~ R(\x, \y) \in \RR.
\end{equation}
The associated loss function is
\begin{equation}
L_R^{\max}(\x, \y) \coloneqq A_R^{\max}(\x) - R(\x, \y).
\label{eq:exact_max_loss}
\end{equation}
The loss is again convex in $R$ and satisfies the property
\begin{equation}
L_R^{\max}(\x, \y) = 0 \iff \y \in \widehat{\y}^{\max}_R(\x).
\end{equation}
The loss is differentiable almost everywhere with gradient 
\begin{equation}
\nabla_\w L^{\max}_{R_\w}(\x, \y) = 
\nabla_\w R_\w(\x, \yhat) 
- \nabla_\w R_\w(\x, \y),
\label{eq:grad_max_loss}
\end{equation}
where $\yhat \coloneqq \mathrm{stop\_gradient}(\widehat{\y}^{\max}_{R_\w}(\x))$.

Unfortunately, this loss, known as the perceptron loss
\citep{collins2002discriminative},
requires an exact maximization oracle, which is intractable when $\cY$ is the
space of sequences. Indeed, the cost of brute force computation
is again $O(V^T)$. 
More generally, Fenchel-Young losses \citep{gfy} provide an elegant way
to construct a convex loss given access to a regularized decoding oracle. 
However, the theory again requires
exact oracles, which are intractable over the space of sequences.

\subsection{Autoregressive models}
\label{sec:background_ar}

An autoregressive (AR) model factorizes the conditional distribution
over sequences as a product of \textbf{token-level} distributions.
We are given a scoring function $q(\s_t, y_t)$,
typically implemented using a causal Transformer,
capturing the affinity between a prefix $\s_t$ and a token $y_t$.
We will also write $q_\w$ to indicate the dependence of $q$ on some parameters
$\w \in \cW$ and we suppose throughout this paper that $q_\w$ is differentiable
in $\w$. We can then define the next-token distribution as
\begin{equation}
\pi_q(y_t|\s_t) 
\coloneqq \frac{\exp(q(\s_t, y_t))}{\sum_{j \in \cV} \exp(q(\s_t, j))}
\end{equation}
and the sequence-level distribution as
\begin{equation}
p^\ar_q(\y|\x) \coloneqq \prod_{t=1}^T \pi_q(y_t | \x \oplus \y_{<t}).
\label{eq:p_ar}
\end{equation}

\paragraph{Token-level log-sum-exp and associated loss.}

The token-level log-partition function is
\begin{equation}
V_q(\s_t)
\coloneqq \underset{y_t \in \cV}{\lse} ~ q(\s_t, y_t)
= \log \sum_{y_t \in \cV} \exp(q(\s_t, y_t)).
\end{equation}

The negative log-likelihood decomposes as a sum of token-level losses:
\begin{equation}
L^\ar_q(\x, \y)
\coloneqq - \log p^\ar_q(\y|\x) 
= -\sum_{t=1}^T \log \pi_q(y_t | \s_t) 
= \sum_{t=1}^T (V_q(\s_t) - q(\s_t, y_t)),
\end{equation}
where $\s_t \coloneqq \x \oplus \y_{<t}$.
The model is using the ground-truth tokens as prefix.  This is known as
\textbf{teacher forcing} \citep{williams1989learning}.  In the imitation
learning literature, teacher forcing corresponds to \textbf{behavior cloning}
\citep{pomerleau1989alvinn, bain1995framework}, where a policy is trained via
supervised learning on expert state-action trajectories.  At inference time,
however, the model conditions on its own previous predictions, not on
ground-truth prefixes. This train-inference mismatch is known as
\textbf{exposure bias} \citep{bengio2015scheduled,ranzato2015sequence} and can
lead to error accumulation.

\paragraph{Analogy with EBMs.}

It is instructive to draw an analogy with EBMs.
We define the ``log-partition'' \textbf{along} the
sequence $\y = (y_1, \dots, y_T)$ as
\begin{align}
A^\ar_q(\x, \y) 
&\coloneqq V_q(\s_1) + V_q(\s_2) + \dots + V_q(\s_T) \\
&=
\underset{y'_1 \in \cV}{\lse} ~ q(\s_1, y'_1) + 
\underset{y'_2 \in \cV}{\lse} ~ q(\s_2, y'_2) + 
\dots +
\underset{y'_T \in \cV}{\lse} ~ q(\s_T, y'_T) \\
&= \log \sum_{\y' \in \cY} \exp(q(\s_1, y'_1) + q(\s_2, y'_2) + \dots 
+ q(\s_T, y'_T)),
\end{align}
where we used the shorthand $\s_t \coloneqq \x \oplus \y_{<t}$.
We also define the cumulative score
\begin{equation}
R_q(\x, \y) 
\coloneqq \sum_{t=1}^T q(\x \oplus \y_{<t}, y_t)
= \sum_{t=1}^T q(\s_t, y_t).
\label{eq:R_q}
\end{equation}
Using these notations, we can equivalently write
\begin{align}
p^\ar_q(\y|\x) &= \exp(R_q(\x, \y) - A^\ar_q(\x, \y)) \label{eq:p_ar_A_ar} \\
L^\ar_q(\x, \y) &= A^\ar_q(\x, \y) - R_q(\x, \y)
\end{align}
by analogy with
\begin{align}
p^\ebm_R(\y|\x) &= \exp(R(\x, \y) - A^\lse_R(\x)) \\
L^\ebm_R(\x, \y) &= A^\lse_R(\x) - R(\x, \y).
\end{align}

\section{Flawed decoding-aware sequence-level loss functions}
\label{sec:flawed}

To address exposure bias, a natural idea is to design sequence-level loss
functions that incorporate the decoding procedure used at inference time. In
this section, we describe two such attempts that appear natural at first sight,
but turn out to be unusable for supervised learning.

\subsection{Greedy decoding based loss function}

In this section, we describe an attempt to define a loss function based on
greedy decoding.

\paragraph{Intractability of exact maximization.}

Even if we assume that $R$ decomposes as \eqref{eq:R_q},
the cost of exact maximization is still $O(V^T)$:
\begin{equation}
\begin{aligned}
A^{\max}_{R_q}(\x)
&\coloneqq \underset{\y \in \cY}{\max} ~ R_q(\x, \y) \\
&= \underset{\y \in \cY}{\max} ~ q(\x, y_1) + q(\x \oplus y_1,y_2) + \dots +
q(\x \oplus \y_{<T}, y_T) \\
&= \underset{y_1 \in \cV}{\max} ~ q(\x, y_1) + \underset{y_2 \in \cV}{\max} ~
q(\x \oplus y_1,y_2) + \dots
+ \underset{y_T \in \cV}{\max} ~ q(\x \oplus \y_{<T}, y_T) \\
&= \underset{y_1 \in \cV}{\max} ~ q(\s_1, y_1) + \underset{y_2 \in \cV}{\max} ~
q(\s_2,y_2) + \dots
+ \underset{y_T \in \cV}{\max} ~ q(\s_T, y_T),
\end{aligned}
\label{eq:A_max_R_q}
\end{equation}
where, with some abuse of notation, $\s_t \coloneqq \x \oplus \y_{<t}$
depends on the variable $\y$ maximized over.

Therefore, a frequently-used inference strategy in AR models is greedy decoding. 

\paragraph{Greedy decoding.}

The key idea is to greedily decode one token at a time given the tokens decoded
so far.  Formally, we define the greedy decoding \textbf{sequence} as
\begin{equation}
\yhat^\greedy_q(\x)
\coloneqq (\wty_1, \dots, \wty_T) \in \cY
\end{equation}
where
\begin{align}
\wty_1 
&\coloneqq \argmax_{y_1 \in \cV} \pi_q(y_1|\tilde{\s}_1)
= \argmax_{y_1 \in \cV} q(\tilde{\s}_1, y_1) \in \cV \\
\wty_2 
&\coloneqq \argmax_{y_2 \in \cV} \pi_q(y_2 | \tilde{\s}_2)
= \argmax_{y_2 \in \cV} q(\tilde{\s}_2, y_2) \in \cV \\
                                        &\vdots \\
\wty_T
&\coloneqq \argmax_{y_T \in \cV} \pi_q(y_T | \tilde{\s}_T) 
= \argmax_{y_T \in \cV} q(\tilde{\s}_T, y_T) \in \cV,
\end{align}
and where we defined the prefix at step $t$ by
\begin{equation}
\tilde{\s}_t \coloneqq \x \oplus \wtyv_{<t}.
\end{equation}

\paragraph{Greedy maximization oracle.}

We define the corresponding greedy maximum \textbf{value} as
\begin{align}
A^\greedy_q(\x) 
& \coloneqq \max_{y_1 \in \cV} q(\tilde{\s}_1, y_1) +
\max_{y_2 \in \cV} q(\tilde{\s}_2, y_2) + 
\dots +
\max_{y_T \in \cV} q(\tilde{\s}_T, y_T) \\
&= q(\tilde{\s}_1, \wty_1) +
q(\tilde{\s}_2, \wty_2) +
\dots +
q(\tilde{\s}_T, \wty_T) \in \RR.
\end{align}
The time complexity is now $O(VT)$, instead of $O(V^T)$ for exact maximization.

\paragraph{Loss definition.}

As an approximation of the exact maximization based loss function in
\eqref{eq:exact_max_loss}, it is a priori natural to define the greedy decoding
loss function
\begin{equation}
L^\greedy_q(\x, \y) \coloneqq A^\greedy_q(\x) - R_q(\x, \y).
\end{equation}
Obviously, we have for all $\x \in \cX$
\begin{equation}
A^{\max}_{R_q}(\x) \ge A^\greedy_q(\x) 
\label{eq:A_max_A_greedy_inequality}
\end{equation}
and therefore for all $(\x, \y) \in \cX \times \cY$
\begin{equation}
L^{\max}_{R_q}(\x, \y) \ge L^\greedy_q(\x, \y).
\end{equation}

\paragraph{Gradient.}

The loss function has discontinuities due to the token-level argmaxes,
but it is differentiable almost everywhere.
\begin{proposition}[Gradient of the greedy-decoding loss]
\label{prop:greedy_gradients}
Suppose $q_\w$ is differentiable in $\w$.
The loss $L^\greedy_{q_\w}(\x, \y)$ is differentiable almost everywhere in $\w$. 
Its gradient is
\begin{equation}
\nabla_\w L^\greedy_{q_\w}(\x, \y) = 
\nabla_\w R_{q_\w}(\x, \yhat) 
- \nabla_\w R_{q_\w}(\x, \y),
\end{equation}
where $\yhat \coloneqq \mathrm{stop\_gradient}(\yhat^\greedy_{q_\w}(\x))$.
\end{proposition}
A proof is given in Appendix \ref{proof:greedy_gradients}.
The gradient has exactly the form we would expect:
it perfectly mirrors the gradient in \eqref{eq:grad_max_loss},
with exact maximization replaced by greedy maximization.

\paragraph{Critical flaws.}

Unfortunately, $L^\greedy_q$ has critical flaws.
\begin{proposition}[Flaws of the greedy decoding based loss]
\label{prop:greedy_loss_unbounded}
The loss function $L^\greedy_q$ is discontinuous, nonconvex in $q$ and unbounded
below.
\end{proposition}
See Appendix~\ref{proof:greedy_loss_unbounded} for a proof.


This loss function can therefore not be used for learning, as the loss would
\textbf{diverge} to $-\infty$.
Constraining the parameters to a bounded set (e.g., via $\ell_2$
regularization) would prevent the loss from diverging, but would not
fix the underlying issue: the constrained minimizer would sit at the boundary,
dominated by the constraint rather than the loss, and the solution would depend
on the arbitrary choice of regularization hyper-parameter.

\paragraph{Relation to violation-fixing updates.}

The unboundedness of $L^\greedy_q$ comes from pairs $(\x, \y)$ for which
$R_q(\x, \yhat^\greedy_q(\x)) < R_q(\x, \y)$, i.e., greedy decoding makes a
search error and the ground truth scores higher than the decoded sequence,
so that $L^\greedy_q(\x, \y) < 0$. In the proof of
Proposition~\ref{prop:greedy_loss_unbounded}, the loss is driven to $-\infty$
precisely by making this search error arbitrarily large.
In the perceptron literature, updates on such pairs are known to break
convergence \citep{huang2012structured}. This motivated early updates
\citep{collins2004incremental}, max-violation updates
\citep{huang2012structured} and beam-search optimization
\citep{wiseman2016sequence}. These approaches compare prefixes only up to a
violation and clip the loss at zero, which makes it bounded below.
They avoid the pitfalls of the greedy loss, but they operate on
unnormalized scores rather than on the autoregressive distribution used for
sampling, produce sparse gradients, and in practice require pre-training
with teacher forcing. 

\subsection{Autoregressive sampling based loss function}

In this section, we describe an attempt to define a loss function based on
autoregressive sampling.

\paragraph{Autoregressive sampling.}

To draw an \iid sequence $Y \sim p^\ar_q(\cdot|\x)$, 
from \eqref{eq:p_ar},
we can use autoregressive sampling (\aka ancestral sampling),
\begin{align}
    Y_1 &\sim \pi_q(\cdot|S_1) \\
    Y_2 &\sim \pi_q(\cdot|S_2) \\
                    &\vdots \\
    Y_T &\sim \pi_q(\cdot|S_T),
\end{align}
where 
$S_t \coloneqq \x \oplus Y_{<t}$,
with
$Y_{<t} \coloneqq (Y_1, \dots, Y_{t-1})$
and
$Y_{<1}$ being an empty sequence.

\paragraph{Intractability of sequence-level log-sum-exp.}

Using the associativity of the log-sum-exp
\citep{mensch2018differentiable,blondel2025autoregressive},
we have that
\begin{align}
A^\lse_{R_q}(\x)
&= \underset{\y \in \cY}{\lse} ~ R_q(\x, \y) \\
&= \underset{\y \in \cY}{\lse} ~ q(\x, y_1) + q(\x \oplus y_1, y_2) + \dots +
q(\x \oplus \y_{<T}, y_T) \\
&= \underset{y_1 \in \cV}{\lse} ~ q(\x, y_1) + \underset{y_2 \in \cV}{\lse} ~
q(\x \oplus y_1,y_2) + \dots
+ \underset{y_T \in \cV}{\lse} ~ q(\x \oplus \y_{<T}, y_T) \\
&= \underset{y_1 \in \cV}{\lse} ~ q(\s_1, y_1) + \underset{y_2 \in \cV}{\lse} ~
q(\s_2,y_2) + \dots
+ \underset{y_T \in \cV}{\lse} ~ q(\s_T, y_T),
\end{align}
where, with some abuse of notation, $\s_t \coloneqq \x \oplus \y_{<t}$
depends on the variable $\y$ being ``log-sum-exped''.

The expression above is the log-sum-exp counterpart of exact maximization in
\eqref{eq:A_max_R_q}. 
Unfortunately, the complexity of brute force is still $O(V^T)$.

\paragraph{Sampling-based cumulant.}

Does there exist a counterpart of greedy maximization for the log-sum-exp?

Recall that $A^\ar_q(\x, \y) = \sum_t V_q(\x \oplus \y_{<t}) = \sum_t V_q(\s_t)$ 
already decomposes the log-partition into token-level log-sum-exps, which are
cheap to compute.  However, unlike $A^\lse_{R_q}(\x)$, it depends on $\y$
through the ground-truth prefix $\s_t = \x \oplus \y_{<t}$, so it cannot serve
as a cumulant on its own.  A natural way to remove this dependency is to replace
the ground-truth prefix with model-sampled prefixes and to average:
\begin{align}
A^\sampling_q(\x) 
&\coloneqq \EE_{Y \sim p^\ar_q(\cdot|\x)} \left[A^\ar_q(\x, Y)\right]
\label{eq:A_sampling} \\
&= \EE_{Y \sim p^\ar_q(\cdot|\x)} \left[
\underset{y_1 \in \cV}{\lse} ~ q(S_1, y_1) + \underset{y_2 \in \cV}{\lse} ~
q(S_2,y_2) + \dots
+ \underset{y_T \in \cV}{\lse} ~ q(S_T, y_T)
\right],
\end{align}
where $S_t \coloneqq \x \oplus Y_{<t}$.

Combining \eqref{eq:A_sampling} and \eqref{eq:p_ar_A_ar},
we can equivalently write
\begin{equation}
A^\sampling_q(\x)
= \EE_{Y \sim p^\ar_q(\cdot|\x)}[R_q(\x, Y) - \log p^\ar_q(Y|\x)].
\label{eq:A_sampling_alt}
\end{equation}
Due to the variational form
\begin{align}
A^\lse_R(\x) 
&= \max_{p \in \cP(\cY)} \EE_{Y \sim p} \left[
R(\x, Y) - \log p(Y)
\right] \\
&= \EE_{Y \sim p^\ebm_R(\cdot|\x)} \left[
R(\x, Y) - \log p^\ebm_R(Y|\x)
\right],
\end{align}
we immediately obtain for all $\x \in \cX$
\begin{equation}
A^\lse_{R_q}(\x) \ge A^\sampling_q(\x),
\end{equation}
which mirrors the inequality \eqref{eq:A_max_A_greedy_inequality} we obtained
for greedy maximization.

\paragraph{Loss definition.}

By analogy with previous sections, we define the sampling based loss
function as
\begin{equation}
L^\sampling_q(\x, \y) \coloneqq A^\sampling_q(\x) - R_q(\x, \y).
\end{equation}
We then have for all $(\x, \y) \in \cX \times \cY$
\begin{equation}
L^\ebm_{R_q}(\x, \y) \ge L^\sampling_q(\x, \y).
\end{equation}

\paragraph{Critical flaws.}

Unfortunately, while the loss function is smooth, it is again not bounded below.
\begin{proposition}[Flaws of the sampling-based loss]
\label{prop:sampling_loss_unbounded}
The sampling based loss $L^\sampling_q$ is nonconvex in $q$ and unbounded below.
\end{proposition}
A proof is given in Appendix \ref{proof:sampling_loss_unbounded}.


This loss function can therefore not be used for learning either,
as the loss would \textbf{diverge} to $-\infty$.

\paragraph{Restriction to finite candidate sets.}

A practical workaround consists in restricting sequence-level losses to a finite
set $\mathcal{U}(\x)$ of sequences decoded by the current model (via beam search
or sampling), taking the max or log-sum-exp over $\mathcal{U}(\x)$ instead
of $\cY$ \citep{edunov2018classical}. 
These losses are bounded below by
construction, because the ground truth $\y$ is replaced with a pseudo-reference
$\u^\star \in \mathcal{U}(\x)$: the candidate closest to the ground truth $\y$
according to a distance $D(\u,\y)$. For example, 
$\log\sum_{\u \in \mathcal{U}(\x)} p(\u|\x) -\log p(\u^\star|\x) \geq 0$.
When $\mathcal{U}(\x)$ consists of $n$
i.i.d.\ samples from the model, maximizing the likelihood of $\u^\star$ can be
viewed as best-of-$n$ distillation
\citep{dong2023raft,zelikman2022star,sessa2025bond}.
However, selecting $\u^\star$ requires a notion of distance $D(\u, \y)$ or 
of reward $R(\x, \u)$, which falls outside the setting we study. 
Note that using the ground-truth $\y$ itself as the target leads to degenerate
models \citep{edunov2018classical}. Finally, the approach is costly, due to the
need to generate $n$ candidate responses.

\section{Autoregressive Gumbel tricks}
\label{sec:autoregressive_gumbel_tricks}

The classical Gumbel trick provides a powerful bridge between maximization and
sampling: sampling from a Gibbs distribution is equivalent to maximizing a
Gumbel perturbed score. This also connects exact maximization to
log-sum-exp. In this section, we develop the autoregressive counterpart of the
classical Gumbel trick, thereby connecting greedy decoding to autoregressive
(ancestral) sampling and unifying the (flawed) loss functions from the previous
section.

\paragraph{Shifted Gumbel noise.}

Following \citet{edpbook},
we consider the shifted standard Gumbel distribution, 
whose probability density function (PDF) is defined by
\begin{equation}
p(x) \coloneqq \exp(-(x + \gamma) - \exp(-(x+\gamma))),
\end{equation}
where $\gamma \approx 0.577 $ is Euler's constant. 
Note that in some publications, the distribution is not shifted by $\gamma$.
Shifting the distribution as we do ensures that the mean is zero.

\paragraph{Gumbel trick for categorical variables.}

Before deriving a Gumbel trick for autoregressive models, we review the standard
Gumbel trick. Since $p^\ebm_R(\cdot|\x) = \softargmax(R(\x, \cdot))$, 
we can apply the standard
Gumbel trick by seeing $\y \in \cY$ as a categorical variable of cardinality
$|\cY|$. Let us define $Z$ such that each $Z(\x, \y)$ is shifted
Gumbel-distributed \iid noise. Then, for any $R$, we have, see e.g. \citep{edpbook},
\begin{equation}
    Y = \ymax_{R + Z}(\x) \implies Y \sim p^\ebm_R(\cdot|\x)
\end{equation}
\begin{equation}
A^\lse_R(\x) = \EE_Z[A^{\max}_{R + Z}(\x)].
\end{equation}

\paragraph{Gumbel trick for sequential variables.}

We now derive analogous results, linking greedy decoding and autoregressive
sampling.
\begin{proposition}[Gumbel trick for autoregressive models]
Let us define $Z$ such that each $Z(\s, y)$ is shifted Gumbel-distributed
\iid noise.  For any $q$, we have
\begin{equation}
Y = \yhat^\greedy_{q + Z}(\x) \implies Y \sim p^\ar_q(\cdot|\x)
\end{equation}
\begin{equation}
A^\sampling_q(\x) = \EE_Z\left[A^\greedy_{q+Z}(\x)\right].
\end{equation}
\label{prop:ancestral_gumbel}
\end{proposition}
A proof is given in Appendix \ref{proof:ancestral_gumbel}.

\paragraph{Unifying the loss functions.}

Since
\begin{align}
L^\sampling_q(\x, \y) &= A^\sampling_q(\x) - R_q(\x, \y) \\
L^\greedy_q(\x, \y) &= A^\greedy_q(\x) - R_q(\x, \y),
\end{align}
Proposition \ref{prop:ancestral_gumbel} immediately implies
\begin{equation}
L^\sampling_q(\x, \y) = \EE_Z\left[L^\greedy_{q+Z}(\x, \y)\right].
\end{equation}
That is, the sampling based loss is the \textbf{expected greedy loss} under
Gumbel perturbations.
This reveals that the two flawed loss functions from the previous section are
in fact two faces of the same coin: greedy decoding is the deterministic
limit, and ancestral sampling is the stochastic counterpart.

\paragraph{Related work.}

Gumbel noise was previously used with AR models \citep{kool2019stochastic} but
the primary goal was to
bridge deterministic beam search and sampling without replacement to create a
practical, low-variance algorithm for generating diverse sequence outputs. 
In contrast, our goal is primarily to bridge
greedy decoding and autoregressive sampling, as well as the resulting (flawed)
loss functions. As we shall see in Section \ref{sec:perspectives}, the Gumbel
trick also plays an important role in justifying reparameterization with a stop
gradient on the greedy decoding.

\section{Moment matching for autoregressive models}
\label{sec:moment_matching}

Since the pathology of the previous loss functions stems from the dependence of
the decoded sequence on the model parameters, we now analyze what happens when
this dependence is removed via a stop gradient.
\subsection{Moment matching directions}
\label{sec:directions}

As we already saw, gradients of the negative log-likelihood of an EBM are given
by
\begin{equation}
\nabla_\w L^\ebm_{R_\w}(\x, \y) 
= \EE_{Y \sim p^\ebm_{R_\w}(\cdot|\x)}\left[\nabla_\w R_\w(\x, Y)\right] -
\nabla_\w R_\w(\x, \y).
\label{eq:ebm_grad}
\end{equation}
Such gradients encourage first-order \textbf{moment matching}.
They are contrasting model samples to ground-truth samples at the 
\textbf{sequence level}.

Gradients of the negative log-likelihood of an AR model are given by
\begin{align}
\nabla_\w L^\ar_{q_\w}(\x, \y) 
&= 
\sum_{t=1}^T
\EE_{Y_t \sim \pi_{q_\w}(\cdot|\s_t)}
\left[
\nabla_\w q_\w(\s_t, Y_t)
\right]
-
\nabla_\w R_{q_\w}(\x, \y) \\
&= 
\sum_{t=1}^T
\EE_{Y_t \sim \pi_{q_\w}(\cdot|\s_t)}
\left[
\nabla_\w q_\w(\s_t, Y_t)
\right]
-
\sum_{t=1}^T \nabla_{\w} q_\w(\s_t, y_t),
\end{align}
where 
$\s_t \coloneqq \x \oplus \y_{<t}$.
Notice that, in the left term, $q_\w$ is using ground-truth tokens as
prefix! This is known as teacher forcing. 
Such gradients are contrasting model samples to ground-truth samples at the 
\textbf{token-level}.

As a middle ground, it is natural to consider the direction
\begin{align}
\d_{q_\w}(\x, \y) 
&\coloneqq \EE_{Y \sim p^\ar_{q_\w}(\cdot|\x)}
\left[
\nabla_\w R_{q_\w}(\x, Y)
\right]
-
\nabla_\w R_{q_\w}(\x, \y) \\
&= \EE_{Y \sim p^\ar_{q_\w}(\cdot|\x)}
\left[
\sum_{t=1}^T
\nabla_\w q_\w(S_t, Y_t)
\right]
-
\sum_{t=1}^T \nabla_{\w} q_\w(\s_t, y_t),
\label{eq:sampling_direction}
\end{align}
where $S_t \coloneqq \x \oplus Y_{<t}$.
Notice that, in the left term, $q_\w$ is now using the model generations as
prefix, not the ground-truth tokens! However, since the model we use for
sampling in the expectation is autoregressive,
not energy-based, it is easy to sample from it.

The direction $\d_{q_\w}(\x, \y)$ can be seen as differentiating \wrt $\w$ the
loss function
\begin{equation}
L_{q_\w}^\mm(\x, \y) \coloneqq 
\EE_{Y \sim p_{q_\w}^\ar(\cdot|\x)}\left[ R_{q_\w}(\x, Y)\right] 
- 
R_{q_\w}(\x, \y),
\end{equation}
while putting a \textbf{stop gradient} on the sampling inside the expectation.

Note that we see from \eqref{eq:A_sampling_alt}
that $L^\mm_q$ differs from $L^\sampling_q$ only by an entropy term,
i.e.,
\begin{equation}
L^\sampling_q(\x, \y) = L_q^\mm(\x, \y) + H(p^\ar_q(\cdot|\x)).
\end{equation}
This entropy term vanishes from the EBM gradient by the \textbf{envelope
theorem} (stationarity of $p^\ebm_R$), but not in the AR case; see
Section~\ref{sec:perspectives} for details.

In fact, we will show that the direction $\d_{q_\w}(\x, \y)$ is a
\textbf{surrogate} gradient:
it is \textbf{not} the gradient of any function, 
but it defines, in function space, a \textbf{descent direction} for the negative
log-likelihood of AR models.

\paragraph{Related work.}

It is informative to compare the direction $\d_{q_\w}$
to scheduled sampling
\citep{bengio2015scheduled,ranzato2015sequence,mihaylova2019scheduled}, which
was introduced to address exposure bias.
In our notation, scheduled sampling flips a coin at each token step $t$ to
decide whether the prefix is extended with a token sampled from the model or
with the ground-truth token:
\begin{equation}
\tilde{Y}_t \coloneqq
\begin{cases}
Y_t \sim \pi_{q_\w}(\cdot|\tilde{S}_t) & \mbox{with probability $\alpha$} \\
y_t & \mbox{with probability $1-\alpha$}
\end{cases},
\qquad
\tilde{S}_t \coloneqq \x \oplus \tilde{Y}_{<t},
\end{equation}
where $\alpha \in [0,1]$ is a hyper-parameter.
The direction used in scheduled sampling then reads
\begin{equation}
\EE_{\tilde{Y}}
\left[
\sum_{t=1}^T
\left(
\EE_{Y_t \sim \pi_{q_\w}(\cdot|\tilde{S}_t)}
\left[
\nabla_\w q_\w(\tilde{S}_t, Y_t)
\right]
-
\nabla_{\w} q_\w(\tilde{S}_t, y_t)
\right)
\right],
\end{equation}
where the mixed prefix $\tilde{S}_t$ is treated as a constant (no gradient is
propagated through the sampling).
When $\alpha = 0$, we have $\tilde{S}_t = \s_t$ and we recover the teacher
forcing gradient $\nabla_\w L^\ar_{q_\w}(\x, \y)$.
When $\alpha = 1$, the prefix is entirely generated by the model.
This leads to potential inconsistencies: the ground-truth token $y_t$ is
the correct next token for the ground-truth prefix $\s_t$, but it is
unlikely to be the correct next token for the mixed prefix $\tilde{S}_t$, 
which contains tokens sampled from the model.
In contrast, the direction $\d_{q_\w}$ is a priori more
principled, because it does not mix sampled prefix with ground-truth tokens.

\subsection{Absence of a corresponding loss}

In this section and the next,
we use a function space representation (\aka tabular setting in RL),
where the function $q$ itself can be seen as the parameter we seek.
In this representation, we write
\begin{equation}
R_q(\x, \y) 
= \sum_{t=1}^T q(\s_t, y_t)
= \langle q, \phi(\x, \y) \rangle.
\end{equation}
Here, $\phi(\x, \y) \coloneqq \sum_{t=1}^T \delta_{(\s_t, y_t)}$
is the feature map that places a unit mass on each state-action pair
$(\s_t, y_t)$ visited along the path defined by $(\x, \y)$.
The moment matching direction, in this function space, then simplifies to
\begin{equation}
\d_q(\x, \y) 
= \EE_{Y \sim p^\ar_q(\cdot|\x)}
\left[
\phi(\x, Y)
\right]
-
\phi(\x, \y).
\label{eq:d_q_functional}
\end{equation}
It is then natural to ask whether there exists a loss function
$L_q(\x, \y)$ whose functional gradient is $\d_q$.

It is well-known that moment matching gradients of the form \eqref{eq:ebm_grad}
are a characteristic of the Gibbs distribution. 
Since $\d_q$ is obtained by replacing $p^\ebm_R$ by $p^\ar_q$,
which is not a Gibbs distribution for $T \ge 2$,
this suggests that $\d_q$ cannot be a gradient.
This is confirmed by the following proposition.
\begin{proposition}[The functional direction $\d_q$ is not the gradient of a loss]
\label{prop:no_loss}
For $T \ge 2$, there does not exist a loss $L_q(\x, \y)$ whose gradient in $q$ 
is the direction $\d_q(\x, \y)$. 
\end{proposition}
A proof is given in Appendix~\ref{proof:no_loss}.

\subsection{Descent direction in function space}

Although $\d_q$ is not the gradient of any loss, we now show that it is, 
in function space, a \textbf{descent direction} for the AR
negative log-likelihood $L^\ar_q$. The key tool is the soft Bellman equation,
which provides a mapping that converts an AR scoring function $q$ into an
equivalent EBM reward function $r$.

\paragraph{AR-to-EBM mapping.}

We define the mapping $r = \cM(q)$ as
\begin{equation}
r(\s_t, y_t) \coloneqq
\begin{cases}
q(\s_t, y_t) &\mbox{if} ~ t = T \\
q(\s_t, y_t) - V_q(\s_t \oplus y_t) &\mbox{if} ~ t < T
\end{cases}.
\end{equation}
We denote by $\partial \cM(q)$ the Jacobian of $\cM$ at $q$ and 
by $\partial \cM(q)^*$ its adjoint.
As proved by \citet{blondel2025autoregressive}, this mapping ensures that
the EBM specified by the reward function $r = \cM(q)$ recovers the AR
distribution exactly: for all $(\x, \y) \in \cX \times \cY$,
\begin{align}
p^\ebm_{R_r}(\y | \x) &= p^\ar_q(\y | \x) \\
A^\lse_{R_r}(\x) &= V_q(\x).
\end{align}

\paragraph{Preconditioned AR gradient.}

Using this mapping, we can relate the moment matching direction $\d_q$ to both
the EBM and AR functional gradients.
\begin{proposition}[Direction as preconditioned gradient]
\label{prop:precond_grad}
For any $q$ and $(\x, \y)$, we have
\begin{equation}
\d_q(\x, \y)  = \left. \nabla_r L^\ebm_{R_r}(\x, \y) \right|_{r=\cM(q)}
\end{equation}
and
\begin{equation}
\partial \cM(q)^* \d_q(\x, \y) = \nabla_q L^{\ar}_q(\x, \y).
\end{equation}
\end{proposition}
A proof is given in Appendix~\ref{proof:precond_grad}.

In other words, $\d_q$ is the EBM functional gradient in the reparameterized
space, and it equals the AR functional gradient preconditioned 
by $(\partial \cM(q)^*)^{-1}$.
Since $\cM$ is invertible \citep{blondel2025autoregressive}, 
$\d_q$ and $\nabla_q L^\ar_q$ point into the same half-space, 
which gives us the descent property.

\begin{proposition}[Descent direction]\label{prop:descent_dir}

With $\d_q$ defined in \eqref{eq:d_q_functional}, we have
\begin{equation}
-\langle \d_q(\x, \y), \nabla_q L^\ar_q(\x, \y) \rangle < 0.
\end{equation}
That is, $-\d_q$ is a descent direction for $L^\ar_q$.
\end{proposition}
A proof is given in Appendix~\ref{proof:descent_dir}.

\subsection{Perspectives}
\label{sec:perspectives}

\subsubsection*{Reparameterization with stop gradient}

The reparameterization trick (path gradient estimator) swaps differentiation and
expectation, but mathematically requires the integrand to be continuous and
differentiable almost everywhere in the parameters \citep{edpbook}.
Therefore, in principle, we cannot use the reparameterization trick for
differentiating 
$L^\sampling_{q_\w}(\x, \y) = \EE_Z[L^\greedy_{q_\w+Z}(\x, \y)]$ 
with respect to $\w$.
However, if we treat the argmax selections in greedy decoding as locally
constant (using a stop gradient), reparameterization yields exactly
$\d_{q_\w}(\x, \y)$.  
Using REINFORCE (score function estimator)
would yield the mathematically-correct gradient of 
$L^\sampling_{q_\w}(\x, \y)$ but we showed that this loss is unbounded below.
In a sense, $\d_{q_\w}(\x, \y)$ is the mathematically-incorrect ``gradient'' of
$\EE_Z[L^\greedy_{q_\w+Z}(\x, \y)]$.

\begin{remark}
We point out that the reparameterization trick (with stop gradient)
at the heart of the moment matching direction $\d_{q_\w}$ does not transfer to
RL post-training.  The direction $\d_{q_\w}$ relies on the fact that the scoring
function $R_{q_\w}$ shares parameters with the policy $p^\ar_{q_\w}$: applying a
stop gradient on the sampling still leaves a nonzero gradient through
$R_{q_\w}$.  In RL, the reward $R$ is usually a separate function,
so stopping the gradient on the sampling zeroes out the signal
entirely. For sequence-level objectives with external rewards, REINFORCE-based
methods therefore remain unavoidable.
\end{remark}

\subsubsection*{The model as its own critic}

We have the following variational inequality for any $R$ and any $q$:
\begin{align}
L^\ebm_R(\x, \y)
&= \max_{p \in \cP(\cY)} \EE_{Y \sim p}[R(\x, Y) - \log p(Y)] - R(\x, \y) \\
&= \EE_{Y \sim p^\ebm_R(\cdot|\x)}[R(\x, Y) - \log p^\ebm_R(Y|\x)] - R(\x, \y) \\
&\ge \EE_{Y \sim p^\ar_q(\cdot|\x)}[R(\x, Y) - \log p^\ar_q(Y|\x)] - R(\x, \y).
\end{align}
The last line can be seen as an \textbf{adversarial loss} where $R$ plays the
role of the \textbf{discriminator} and $p^\ar_q$ plays the role of the
\textbf{generator}. In the RL terminology, these correspond to the concepts of
\textbf{critic} and \textbf{actor}, respectively.

In particular, by setting $R=R_{q_\w}$ and $q = q_\w$, we obtain the loss
function
\begin{equation}
L_{q_\w}(\x, \y)
\coloneqq
\EE_{Y \sim p^\ar_{q_\w}(\cdot|\x)}[R_{q_\w}(\x, Y) - \log p^\ar_{q_\w}(Y|\x)] 
- R_{q_\w}(\x, \y).
\end{equation}
If we put a stop gradient on $p^\ar_{q_\w}$, we immediately obtain that the
gradient in $\w$ recovers the moment matching direction $\d_{q_\w}$.  In other
words, $\d_{q_\w}$ corresponds to using the same underlying model for the
discriminator (reward model) and for the generator, with a stop gradient on the
generator!

\subsubsection*{Non-stationarity gap between EBMs and AR models}

In this section, we drop $\x$ for notational simplicity.

Let us define
\begin{equation}
F(p, R) 
\coloneqq \langle p, R \rangle - \langle p, \log p \rangle
= \EE_{Y \sim p}[R(Y) - \log p(Y)].
\end{equation}
For any mapping $R \mapsto p_R$, we have by the chain rule
\begin{align}
\nabla_R F(p_R, R)
&= (\partial_R p_R)^\top \nabla_1 F(p_R, R) + \nabla_2 F(p_R, R) \\
&= (\partial_R p_R)^* \nabla_1 F(p_R, R) + p_R.
\end{align}
Observe that
\begin{equation}
A^\lse_R = F(p^\ebm_R, R).
\end{equation}
Since $p^\ebm_R$ maximizes $F(\cdot, R)$, we have
$\nabla_1 F(p^\ebm_R, R) = 0$ and therefore
\begin{equation}
\nabla_R A^\lse_R = p^\ebm_R.
\end{equation}
This is a consequence of the \textbf{envelope theorem} \citep{edpbook}.

On the other hand, from \eqref{eq:A_sampling_alt},
\begin{equation}
A^\sampling_q = F(p^\ar_q, R_q).
\end{equation}
Since $p^\ar_q$ does \textbf{not} maximize $F(\cdot, R_q)$
over all distributions,
$\nabla_1 F(p^\ar_q, R_q) \neq 0$ and
\begin{equation}
\nabla_q A^\sampling_q =
\underbrace{(\partial_q p^\ar_q)^* \nabla_1 F(p^\ar_q,
R_q)}_{\text{non-stationarity term}} + p^\ar_q.
\end{equation}
The direction
\begin{equation}
\d_q(\y) 
= \nabla_q A^\sampling_q - \phi(\y)
= p^\ar_q - \phi(\y) 
\end{equation}
therefore corresponds to
\textbf{dropping the non-stationarity term}, i.e., treating $p^\ar_q$ \textbf{as
if} it satisfied the stationarity condition (as the EBM does).
Here, $\phi(\y) = \delta_\y$ is the Dirac distribution centered at the ground
truth, so $\d_q$ matches the model distribution $p^\ar_q$ to $\delta_\y$.
This is analogous to the stop-gradient trick: we differentiate through $R_q$
but treat the dependence of $p^\ar_q$ on $q$ as fixed.

\subsection{Limitations}

\subsubsection*{Function space vs.\ parameter space}

We emphasize that Proposition~\ref{prop:descent_dir} holds in function space.
In parameter space with $q = q_\w$, let $J(\w) \coloneqq \partial q_\w /
\partial \w$ denote the Jacobian of the parameterization.  By the chain rule,
the parameter-space direction is $\d_{q_\w} = J(\w)^* \d_q$ and the parameter
gradient is $\nabla_\w L^\ar_{q_\w} = J(\w)^* \nabla_q L^\ar_q$.  Their inner
product equals $\langle \d_q, J(\w) J(\w)^* \nabla_q L^\ar_q \rangle$.
Because the Gram operator $J(\w) J(\w)^*$ deforms inner products, descent in
function space does not imply descent in parameter space in general, even under
a linear parameterization $q_\w = \mathbf{\Phi} \w$.
That being said, in deep learning, descent guarantees are routinely missing
(Adam, momentum, etc.).
The perspectives in Section~\ref{sec:perspectives} provide some
motivation for the moment matching direction, beyond the descent property.

\subsubsection*{Variance}

Although $\d_{q_\w}$ avoids the pathologies of the flawed losses from Section
\ref{sec:flawed},
its practical use is limited by \textbf{high variance}.
The teacher forcing gradient $\nabla_\w L^\ar_{q_\w}$ decomposes as a sum of
$T$ token-level expectations, each over $\cV$ (of size $V$), which can be
computed exactly via a single forward pass (no sampling is needed).
In contrast, the moment matching direction $\d_{q_\w}$ involves an expectation
over $p^\ar_{q_\w}$, a distribution over $\cY = \cV^T$, which can only be
estimated by Monte Carlo sampling.  Since sampled prefixes $S_t = \x \oplus
Y_{<t}$ are correlated across time steps, the variance of this estimator grows
with the sequence length $T$.
Since teacher forcing provides the exact gradient 
$\nabla_\w L^\ar_{q_\w}$ cheaply (a single forward pass, no sampling) anyway, 
teacher forcing remains preferable in practice.

\section{Conclusion}

In this paper, 
we studied the possibility to design decoding-aware
sequence-level loss functions for addressing exposure bias in
supervised learning of autoregressive (AR) models.

We first studied $L^\greedy_q$, a loss function based on greedy decoding.
We showed that it is discontinuous and unbounded below.
We then studied $L^\sampling_q$, a loss function based on autoregressive
sampling. While the loss function is smooth, it remains unbounded below.
Interestingly, we showed that the two loss functions are related through
$L^\sampling_q = \EE_Z[L^\greedy_{q+Z}]$, where $Z$ is \iid Gumbel noise.
Unfortunately, neither loss function can be used for learning, 
as the loss would diverge to $-\infty$.

Since the pathology of these loss functions stems from the dependence of the
decoded sequence on the model parameters, we analyzed the update direction
obtained by applying a stop gradient on the sampling procedure, leading to the
moment matching direction $\d_q$.
We saw that this direction corresponds to dropping the
problematic non-stationarity term in $\nabla_q A^\sampling_q$.

Nevertheless, while the moment matching direction is theoretically appealing,
its practical use is limited by high variance.
Since $-\d_q$ is merely a descent direction for $L^\ar_q$ in function space,
while teacher forcing provides the exact gradient 
$\nabla_q L^\ar_q$ cheaply anyway (a single forward pass, no sampling),
teacher forcing remains preferable in practice.

We conclude that teacher forcing may well be ``all you need'' for the supervised
learning phase of AR models from static datasets; see also
\citep{foster2024behavior, blondel2025autoregressive} for additional insights. 
While the exposure bias concern is real in principle, our paper suggests that
it cannot be resolved through sequence-level loss functions obtained by
plugging a greedy or sampling decoding oracle into a cumulant, without
introducing severe pathologies. Violation-based margin losses
\citep{collins2004incremental,huang2012structured,wiseman2016sequence}
avoid these pathologies, but at the cost of sparse gradients and a departure
from the probabilistic model, which is why they are used on top of, rather
than instead of, teacher forcing.

If one wishes to directly address exposure bias, it is safer to change the
training data distribution rather than the loss function. For instance, online
policy distillation \citep{agarwal2024policy} resolves the train-inference
mismatch by querying a teacher model on prefixes generated by the student, but
crucially, it still relies on the mathematically well-behaved token-level
cross-entropy loss. Alternatively, if true sequence-level signals are desired, 
they are best incorporated via RL-based post-training. RL bypasses the 
aforementioned pathologies thanks to the score function estimator (REINFORCE), 
allowing the use of a task-specific fixed reward $R(\x, \y)$ or a distance
metric $R(\x, \y) = -D(\y, \y^\star(\x))$ to the ground-truth $\y^\star(\x)$
\citep{kreutzer2017bandit}.

\bibliography{main}

\newpage

\appendix

\section*{Appendix}

\section{Proofs}

\subsection{Greedy-decoding loss gradients (Proof of
Proposition~\ref{prop:greedy_gradients})}
\label{proof:greedy_gradients}

Recall that $L^\greedy_{q_\w}(\x, \y) = A^\greedy_{q_\w}(\x) - R_{q_\w}(\x, \y)$,
where $A^\greedy_{q_\w}(\x) = R_{q_\w}(\x, \yhat^\greedy_{q_\w}(\x))$.

\paragraph{Differentiability almost everywhere.}

Let $\w$ be such that none of the step-wise greedy choices
have ties, i.e.,
$q_\w(\tilde{\s}_t, y_t^{(1)}) \neq q_\w(\tilde{\s}_t, y_t^{(2)})$
for any step $t \in [T]$ and any distinct tokens $y_t^{(1)}, y_t^{(2)} \in \cV$,
where $\tilde{\s}_t \coloneqq \x \oplus \wtyv_{<t}$.
There exists a neighborhood $U$ of $\w$ such that any $\w' \in U$ maintains the
same strict inequalities at each step.
Consequently, $\yhat^\greedy_{q_{\w'}}(\x) = \yhat^\greedy_{q_\w}(\x)$
for all $\w' \in U$, so $\w \mapsto \yhat^\greedy_{q_\w}(\x)$ is locally
constant and its Jacobian is $\mathbf{0}$.
The set of $\w$ where ties occur has Lebesgue measure zero,
so the loss is differentiable almost everywhere.

\paragraph{Gradient computation.}

At any point of differentiability, $\yhat \coloneqq \yhat^\greedy_{q_\w}(\x)$
is locally constant in $\w$. By the chain rule,
\begin{align}
\nabla_\w L^\greedy_{q_\w}(\x, \y) 
&= \nabla_\w \left[ R_{q_\w}(\x, \yhat) - R_{q_\w}(\x, \y) \right] \\
&= \nabla_\w R_{q_\w}(\x, \yhat) - \nabla_\w R_{q_\w}(\x, \y),
\end{align}
where $\yhat$ is treated as a constant (stop gradient).

\subsection{Discontinuity, non-convexity, and unboundedness of greedy decoding loss
(Proof of Proposition~\ref{prop:greedy_loss_unbounded})}
\label{proof:greedy_loss_unbounded}

Set $T = 2$, $V = 2$ and $\cV = \{1, 2\}$ for simplicity. 
Consider w.l.o.g.\ $y_1 = y_2 = 1$.
The greedy loss is
\begin{align*}
L^\greedy_q(\x, \y)
&= A^\greedy_q(\x) - R_q(\x, \y) \\
&= \max_{y'_1} q(\s_1, y'_1) - q(\s_1, y_1) 
+ \max_{y'_2} q(\x \oplus \wty_1, y'_2) - q(\s_2, y_2),
\end{align*}
where $\wty_1 = \argmax_{y'_1} q(\s_1, y'_1)$.

\paragraph{Discontinuity and non-convexity.}
Define the family $q_\epsilon$ by
\begin{equation}
q_\epsilon(\s_1, 1) = 0, \quad q_\epsilon(\s_1, 2) = \epsilon, \quad
q_\epsilon(\x \oplus 1, \cdot) = (0, 1), \quad
q_\epsilon(\x \oplus 2, \cdot) = (0, 2).
\end{equation}
Then $\lim_{\epsilon \to 0^-} q_\epsilon = \lim_{\epsilon \to 0^+} q_\epsilon$, yet
\begin{equation}
\lim_{\epsilon \to 0^-} L^\greedy_{q_\epsilon}(\x, \y) = 1
\neq 2
= \lim_{\epsilon \to 0^+} L^\greedy_{q_\epsilon}(\x, \y),
\end{equation}
since for $\epsilon < 0$ the greedy token is $\wty_1 = 1$ (giving loss $0 + 1 = 1$),
while for $\epsilon > 0$ it switches to $\wty_1 = 2$ (giving loss $\epsilon + 2 \to 2$).
Hence $L^\greedy_q$ is not continuous in $q$.
Any convex function is locally Lipschitz continuous so the function $L^\greedy_q$ cannot be convex.

\paragraph{Unboundedness.}
Define $q_\alpha$ by
\begin{equation}
q_\alpha(\s_1, 1) = 0, \quad q_\alpha(\s_1, 2) = 1, \quad
q_\alpha(\x \oplus 1, \cdot) = (\alpha, 0), \quad
q_\alpha(\x \oplus 2, \cdot) = (0, 0).
\end{equation}
Then $\wty_1 = 2$ and
\begin{equation}
L^\greedy_{q_\alpha}(\x, \y) = 1 - \alpha 
\underset{\alpha \to +\infty}{\longrightarrow} -\infty.
\end{equation}
Hence the greedy decoding loss is unbounded below.
Note also that if we set $\alpha = 2|\sin(\beta)|$ for instance,
we see that the loss has infinitely many zeros in $\beta$.

\subsection{Nonconvexity and unboundedness of sampling based loss
(Proof of Proposition~\ref{prop:sampling_loss_unbounded})}
\label{proof:sampling_loss_unbounded}


\paragraph{Non-convexity.} 
We demonstrate that $L^\sampling_q$ is not convex in $q$ by constructing a counterexample that violates Jensen's inequality. 
Set $T=2$, $V=2$, and $\cV=\{1, 2\}$. Consider w.l.o.g.\ $y_1=y_2=1$. 
Because the reward term $R_q(\x, \y)$ is linear in $q$, the loss $L^\sampling_q$ is convex if and only if the expected greedy value $f(q) \coloneqq \EE_Z \left[A^\greedy_{q+Z}(\x)\right]$ is convex. 
Using Proposition~\ref{prop:ancestral_gumbel}, we have the closed-form expression:
\begin{equation}
    f(q) = \mathrm{LSE}_{y_1 \in \cV} \, q(\s_1, y_1) + \sum_{y_1 \in \cV} \pi_q(y_1 | \s_1) \mathrm{LSE}_{y_2 \in \cV} \, q(\x \oplus y_1, y_2),
\end{equation}
where $\pi_q(y_1|\s_1) \propto \exp(q(\s_1, y_1))$. 

Define two configurations $q_1$ and $q_2$ parameterized by $\alpha, \beta > 0$:
\begin{align*}
    q_1(\s_1, \cdot) &= (\alpha, -\alpha), & q_1(\x \oplus 1, \cdot) &= (0, 0), & q_1(\x \oplus 2, \cdot) &= (0, \beta) \\
    q_2(\s_1, \cdot) &= (-\alpha, \alpha), & q_2(\x \oplus 1, \cdot) &= (0, 0), & q_2(\x \oplus 2, \cdot) &= (0, 0).
\end{align*}
Their midpoint $q_M = \frac{1}{2}(q_1 + q_2)$ is given by:
\begin{equation}
    q_M(\s_1, \cdot) = (0, 0), \quad q_M(\x \oplus 1, \cdot) = (0, 0), \quad q_M(\x \oplus 2, \cdot) = (0, \textstyle\frac{\beta}{2}).
\end{equation}
Evaluating $f$ for $q_1$ and $q_2$, we obtain:
\begin{align*}
    f(q_1) &= \log(e^\alpha + e^{-\alpha}) + \pi_{q_1}(1|\s_1)\log(2) + \pi_{q_1}(2|\s_1)\log(1+e^\beta) \\
    f(q_2) &= \log(e^{-\alpha} + e^\alpha) + \pi_{q_2}(1|\s_1)\log(2) + \pi_{q_2}(2|\s_1)\log(2).
\end{align*}
As $\alpha \to +\infty$, the routing probabilities become deterministic: $\pi_{q_1}(1|\s_1) \to 1$ and $\pi_{q_2}(2|\s_1) \to 1$. Consequently, both $f(q_1)$ and $f(q_2)$ scale asymptotically as $\alpha + \log(2)$. Thus, their convex combination is bounded:
\begin{equation}
    \lim_{\alpha \to +\infty} \left[ \frac{1}{2}f(q_1) + \frac{1}{2}f(q_2) - \alpha \right] = \log(2).
\end{equation}
However, for the midpoint $q_M$, the first-step logits are zero, giving uniform routing probabilities $\pi_{q_M}(\cdot|\s_1) = \frac{1}{2}$. Evaluating $f(q_M)$ yields:
\begin{equation}
    f(q_M) = \log(2) + \frac{1}{2}\log(2) + \frac{1}{2}\log(1+e^{\beta/2}).
\end{equation}
For any arbitrarily large $\alpha$, we can choose $\beta \gg \alpha$ such that $f(q_M) > \frac{1}{2}f(q_1) + \frac{1}{2}f(q_2)$, which strictly violates the necessary condition for convexity. Therefore, $L^\sampling_q$ is not convex.

\paragraph{Unboundedness.}
Set $T = 2$, $V = 2$ and $\cV = \{1, 2\}$ for simplicity.
Consider w.l.o.g.\ $y_1 = y_2 = 1$.
Using Proposition~\ref{prop:ancestral_gumbel}, we have
\begin{equation}
L^\sampling_q(\x, \y) = 
\EE_Z\left[A^\greedy_{q + Z}(\x) \right] - R_q(\x, \y).
\end{equation}
Define $q_\alpha$ by
\begin{equation}
q_\alpha(\s_1, 1) = 0, \quad q_\alpha(\s_1, 2) = 0, \quad
q_\alpha(\x \oplus 1, \cdot) = (\alpha, 0), \quad
q_\alpha(\x \oplus 2, \cdot) = (0, 0).
\end{equation}
Then $R_{q_\alpha}(\x, \y) = q_\alpha(\s_1, 1) + q_\alpha(\x \oplus 1, 1) = \alpha$.
The greedy value under shifted Gumbel noise $Z$ is
\begin{align*}
A^\greedy_{q_\alpha + Z}(\x)
&= \max_{y'_1} Z(\s_1, y'_1)
+ \max_{y'_2} \big(q_\alpha(\x \oplus \wty_1, y'_2) + Z(\x \oplus \wty_1, y'_2)\big),
\end{align*}
where $\wty_1 = \argmax_{y'_1} Z(\s_1, y'_1)$.
Taking expectations, the first term gives a constant \\
$c \coloneqq \EE_Z[\max(Z(\s_1, 1), Z(\s_1, 2))]$.
Since $q_\alpha(\s_1, 1) = q_\alpha(\s_1, 2) = 0$,
we have $\wty_1 = 1$ or $2$ with equal probability $\frac12$.
By the law of total expectation,
\begin{align*}
\EE_Z\!\left[\max_{y'_2}\! (q_\alpha\! +\! Z)(\x \oplus \wty_1, y'_2)\right]
&= \tfrac12\, \EE_Z\!\left[\max(\alpha + Z(\x \oplus 1, 1),\, Z(\x \oplus 1, 2))\right]
+ \tfrac12\, c.
\end{align*}
Recall from Section~\ref{sec:autoregressive_gumbel_tricks} that $Z$ is shifted
Gumbel noise, whose CDF is $\exp\!\big({-e^{-(\zeta + \gamma)}}\big)$,
where $\gamma$ is Euler's constant.
By independence, the CDF of $\max(\alpha + Z(\x \oplus 1, 1),\, Z(\x \oplus 1, 2))$
is $F(\zeta) = \exp\!\big({-e^{-(\zeta + \gamma)}(e^\alpha + 1)}\big)
= \exp\!\big({-e^{\log(e^\alpha + 1) - \gamma - \zeta}}\big)$,
which is Gumbel with location $\log(1 + e^\alpha) - \gamma$ and scale $1$.
Its expectation is therefore $\log(1 + e^\alpha)$.
The same computation with $\alpha = 0$ gives $c = \log 2$.
Therefore,
\begin{align*}
L^\sampling_{q_\alpha}(\x, \y)
&= \tfrac32\, c + \tfrac12 \log(1 + e^\alpha) - \alpha
= \tfrac32 \log 2 + \tfrac12 \log(1 + e^\alpha) - \alpha
\underset{\alpha \to +\infty}{\longrightarrow} -\infty.
\end{align*}

\subsection{Proof of autoregressive Gumbel tricks (Proposition~\ref{prop:ancestral_gumbel})}
\label{proof:ancestral_gumbel}

Recall that $\yhat^\greedy_q(\x) = (\wty_1, \dots, \wty_T)$ is defined as
\begin{align}
\wty_1 &\coloneqq \argmax_{y_1 \in \cV} q(\tilde{\s}_1, y_1) \in \cV \\
\wty_2 &\coloneqq \argmax_{y_2 \in \cV}
q(\tilde{\s}_2, y_2) \in \cV, \\
&\vdots \\
\wty_T &\coloneqq \argmax_{y_T \in \cV}
q(\tilde{\s}_T, y_T) \in \cV,
\end{align}
where $\tilde{\s}_t \coloneqq \x \oplus \wtyv_{<t}$.

\paragraph{Ancestral sampling.}

By successive applications of the Gumbel trick for categorical variables,
\begin{align}
Y_1 
&= \argmax_{y_1 \in \cV} (q + Z)(S_1, y_1)
\sim \pi_q(\cdot|S_1) \\
Y_2 
&= \argmax_{y_2 \in \cV} (q + Z)(S_2, y_2)
\sim \pi_q(\cdot|S_2) \\
&\vdots \\
Y_T 
&= \argmax_{y_T \in \cV} (q + Z)(S_T, y_T)
\sim \pi_q(\cdot|S_T),
\end{align}
where $S_t$ now uses the perturbed greedy prefix.

\paragraph{Cumulant.}

We then have
\begin{align*}
A^\greedy_{q+Z}(\x) 
= \max_{y_1 \in \cV} (q+Z)(S_1, y_1) + \max_{y_2 \in \cV} (q+Z)(S_2, y_2)  +
\ldots + \max_{y_T \in \cV} (q+Z)(S_T, y_T).
\end{align*}
Therefore
\begin{align*}
A^\sampling_q(\x) 
& = \EE_Z\left[ A^\greedy_{q+Z}(\x) \right] \\
& = \EE_{Z_1} \max_{y_1 \in \cV} ~ 
(q + Z)(S_1, y_1) \\
& \quad + \EE_{Z_1} \EE_{Z_2} \max_{y_2 \in \cV} ~ 
(q + Z)(S_2, y_2) \\
& \quad + \ldots \\
& \quad + \EE_{Z_1} \ldots \EE_{Z_T} \max_{y_T \in \cV} ~
(q + Z)(S_T, y_T) \\
& = \underset{y_1 \in \cV}{\lse} ~ q(S_1, y_1) \\
& \quad + \EE_{Z_1} \underset{y_2 \in \cV}{\lse} ~ q(S_2, y_2) \\
& \quad + \ldots \\
& \quad +  \EE_{Z_1} \ldots \EE_{Z_{T-1}} \underset{y_T \in \cV}{\lse} ~
q(S_T, y_T) \\
& = \EE_{Y \sim p^\ar_q(\cdot|\x)}\left[\sum_{t=1}^T \underset{y_t \in \cV}{\lse} ~
q(S_t, y_t)\right] \\
&= \EE_{Y \sim p^\ar_q(\cdot|\x)}\left[A^\ar_q(\x, Y)\right].
\end{align*}

\subsection{The direction is not a gradient (Proof of Proposition
\ref{prop:no_loss})}
\label{proof:no_loss}

If there existed a loss $L_q(\x, \y)$ such that
$\nabla_q L_q(\x, \y) = \d_q(\x, \y)$,
then the Jacobian $\partial \d_q(\x, \y)$
would be self-adjoint.
Since $\d_q(\x, \y) = \mu(q, \x) - \phi(\x, \y)$,
where $\mu(q, \x) \coloneqq \EE_{Y \sim p^\ar_q(\cdot|\x)}[\phi(\x, Y)]$,
the Jacobian $\partial_1 \mu(q, \x)$ would also be self-adjoint
(that is, if we materialized the Jacobian as a matrix, it would be
symmetric).

However, the step-1 component
$[\mu(q, \x)](\s_1, y_1) = \pi_q(y_1|\s_1)$
depends only on $\{q(\s_1, y) : y \in \cV\}$, so
\begin{equation}
[\partial_1 \mu(q, \x) \cdot h](\s_1, y_1) = 0
\end{equation}
for any perturbation $h$ supported on steps $t \ge 2$.
On the other hand, for $t \ge 2$,
the component $[\mu(q, \x)](\s_t, y_t)$ involves
$\pi_q(y_1|\s_1) \cdots \pi_q(y_t|\s_t)$ and therefore depends on
$q(\s_1, \cdot)$, so
$[\partial_1 \mu(q, \x) \cdot g](\s_t, y_t) \neq 0$
for generic perturbations $g$ supported at step~1.
The Jacobian is therefore not self-adjoint, a contradiction.

\subsection{Direction as preconditioned gradient 
(Proof of Proposition~\ref{prop:precond_grad})}
\label{proof:precond_grad}

\paragraph{First claim.}

For any $r$ and $(\x, \y)$, we have
\begin{equation}
\nabla_r A^\lse_{R_r}(\x) 
= \EE_{Y \sim p^\ebm_{R_r}(\cdot|\x)}[\phi(\x, Y)]. 
\end{equation}
Therefore,
\begin{align*}
\left. \nabla_r L^\ebm_{R_r}(\x, \y) \right|_{r=\cM(q)}
& = \EE_{Y \sim p^\ebm_{R_r}(\cdot|\x)}[\phi(\x, Y)] - \phi(\x, \y) \\
& = \EE_{Y \sim p^\ar_q(\cdot|\x)}[\phi(\x, Y)] - \phi(\x, \y) \\
& = \d_q(\x, \y),
\end{align*}
where the second line uses $p^\ebm_{R_r}(\cdot|\x) = p^\ar_q(\cdot|\x)$
as proved by \citet{blondel2025autoregressive}.

\paragraph{Second claim.}

Since $L^\ebm_{R_{\cM(q)}}(\x, \y) = L^\ar_q(\x, \y)$, we get by the chain rule
\begin{align*}
\nabla_q L^\ar_q(\x, \y) 
= \partial \cM(q)^* \left. \nabla_r L^\ebm_{R_r}(\x, \y) \right|_{r=\cM(q)}
= \partial \cM(q)^* \d_q(\x, \y).
\end{align*}

\subsection{Descent direction 
(Proof of Proposition~\ref{prop:descent_dir})}
\label{proof:descent_dir}

We decompose any function $h$ on state-action pairs by step:
$h = (h_1, \ldots, h_T)$, where $h_t$ is the restriction of $h$ to
step $t$.

\paragraph{Jacobian of $\cM$.}

Since $[\cM(q)](\s_t, y_t) = q(\s_t, y_t) - V_q(\s_t \oplus y_t)$ for $t < T$
and $[\cM(q)](\s_T, y_T) = q(\s_T, y_T)$, the Jacobian of $\cM$ at $q$ acts on
a perturbation $h$ as
\begin{equation}
[\partial \cM(q) \cdot h](\s_t, y_t) = 
\begin{cases}
h(\s_t, y_t) - [J_{t+1} h_{t+1}](\s_t, y_t) & \text{if } t < T \\
h(\s_t, y_t) & \text{if } t = T,
\end{cases}
\end{equation}
where the linear operator $J_{t+1}$ is defined by
\begin{equation}
[J_{t+1} h_{t+1}](\s_t, y_t) 
\coloneqq \EE_{Y_{t+1} \sim \pi_q(\cdot|\s_{t+1})}
[h_{t+1}(\s_{t+1}, Y_{t+1})], \quad \text{with } \s_{t+1} \coloneqq \s_t \oplus y_t.
\end{equation}

Indeed, this follows from a Taylor expansion of $V_q$.
Since $V_q(\s_{t+1}) = \log \sum_{y \in \cV} \exp(q(\s_{t+1}, y))$,
we have
\begin{align*}
V_{q + \epsilon h}(\s_{t+1})
&= \log \sum_{y \in \cV} \exp\big(q(\s_{t+1}, y) + \epsilon\, h(\s_{t+1}, y)\big) \\
&= \log \sum_{y \in \cV} \exp(q(\s_{t+1}, y))
\big(1 + \epsilon\, h(\s_{t+1}, y) + o(\epsilon)\big) \\
&= \log \left[\left(\sum_{y \in \cV} \exp(q(\s_{t+1}, y))\right)
\left(1 + \epsilon \sum_{y \in \cV} \pi_q(y|\s_{t+1})\, h(\s_{t+1}, y)
+ o(\epsilon)\right)\right] \\
&= V_q(\s_{t+1}) 
+ \log\!\left(1 + \epsilon \, \EE_{Y_{t+1} \sim \pi_q(\cdot|\s_{t+1})}
[h_{t+1}(\s_{t+1}, Y_{t+1})] + o(\epsilon)\right) \\
&= V_q(\s_{t+1}) 
+ \epsilon \, \EE_{Y_{t+1} \sim \pi_q(\cdot|\s_{t+1})}
[h_{t+1}(\s_{t+1}, Y_{t+1})] 
+ o(\epsilon),
\end{align*}
where we used $\exp(a + \epsilon b) = \exp(a)(1 + \epsilon b + o(\epsilon))$
and $\log(1 + u) = u + o(u)$.

\paragraph{Contraction property.}

We claim that $\|J_{t+1} h_{t+1}\| \leq \|h_{t+1}\|$.
Indeed, by Jensen's inequality (the square is convex),
\begin{align*}
\|J_{t+1} h_{t+1}\|^2
&= \sum_{\s_t, y_t}
\Big(\EE_{Y_{t+1} \sim \pi_q(\cdot|\s_{t+1})}
[h_{t+1}(\s_{t+1}, Y_{t+1})]\Big)^2 \\
&\leq \sum_{\s_t, y_t} \sum_{y_{t+1}}
\pi_q(y_{t+1}|\s_{t+1})\, h_{t+1}(\s_{t+1}, y_{t+1})^2.
\end{align*}
Since the map $(\s_t, y_t) \mapsto \s_{t+1}$ is a bijection
and each weight satisfies $\pi_q(y_{t+1}|\s_{t+1}) \leq 1$,
we obtain
$\|J_{t+1} h_{t+1}\|^2
\leq \sum_{\s_{t+1}, y_{t+1}} h_{t+1}(\s_{t+1}, y_{t+1})^2
= \|h_{t+1}\|^2$.

\paragraph{Positive definiteness.}

We have
\begin{align*}
\langle h, \partial \cM(q) \cdot h \rangle
= \sum_{t=1}^T \|h_t\|^2
- \sum_{t=2}^T \langle h_{t-1}, J_t h_t \rangle.
\end{align*}
By Cauchy--Schwarz and the contraction property,
\begin{equation}
\sum_{t=2}^T \langle h_{t-1}, J_t h_t \rangle
\leq \sum_{t=2}^T \|h_{t-1}\| \|h_t\|.
\end{equation}
Since for any real numbers $a, b$ we have $ab \leq \frac{1}{2}(a^2 + b^2)$,
\begin{align*}
\sum_{t=2}^T \|h_{t-1}\| \|h_t\|
&\leq \frac{1}{2} \sum_{t=2}^T 
(\|h_{t-1}\|^2 + \|h_t\|^2) \\
&= \frac{1}{2}\|h_1\|^2 
+ \sum_{t=2}^{T-1} \|h_t\|^2
+ \frac{1}{2}\|h_T\|^2
\leq \sum_{t=1}^T \|h_t\|^2.
\end{align*}
Moreover, $ab = \frac{1}{2}(a^2 + b^2)$ holds if and only if $a = b$. Thus,
equality in the first step above holds iff $\|h_t\| = \alpha$ for all $t$, in
which case
$\sum_{t=2}^T \|h_{t-1}\| \|h_t\| = (T-1)\alpha^2 < T\alpha^2 
= \sum_{t=1}^T \|h_t\|^2$ whenever $\alpha \neq 0$.
Therefore $\langle h, \partial \cM(q) \cdot h \rangle > 0$
for all $h \neq 0$.

\paragraph{Conclusion.}

Using Proposition~\ref{prop:precond_grad}:
\begin{align*}
\langle -\d_q(\x, \y), \nabla_q L^\ar_q(\x, \y) \rangle 
= -\langle \d_q(\x, \y), \partial \cM(q)^* \d_q(\x, \y) \rangle
< 0,
\end{align*}
for any $q, \x, \y$ such that $\d_q(\x, \y)\neq 0$.

\end{document}